%% file: egpaper_final.tex
\documentclass[10pt,twocolumn,letterpaper]{article}

\usepackage{cvpr}
\usepackage{times}
\usepackage{epsfig}
\usepackage{graphicx}
\usepackage{amsmath}
\usepackage{amssymb}
\usepackage{times}  % DO NOT CHANGE THIS
\usepackage{helvet} % DO NOT CHANGE THIS
\usepackage{courier}  % DO NOT CHANGE THIS
\usepackage[hyphens]{url}  % DO NOT CHANGE THIS
\usepackage{graphicx} % DO NOT CHANGE THIS
\usepackage{booktabs}
\usepackage{graphicx}  % DO NOT CHANGE THIS
\usepackage{pifont}
\usepackage{wasysym}
\usepackage{amssymb}
\usepackage{color}
\usepackage{multirow}
\usepackage{subcaption}
\usepackage{authblk}

\newcommand{\pxj}[1]{\textcolor[rgb]{0,0,0}{#1}}

\usepackage[pagebackref=true,breaklinks=true,letterpaper=true,colorlinks,bookmarks=false]{hyperref}

\cvprfinalcopy % *** Uncomment this line for the final submission

\def\cvprPaperID{****} % *** Enter the CVPR Paper ID here

\ifcvprfinal\fi
\begin{document}

%%%%%%%%% TITLE
\title{Suppressing Uncertainties for Large-Scale Facial Expression Recognition}

\author[1,2]{Kai Wang$^*$}
\author[1]{Xiaojiang Peng\thanks{Equally-contributed first authors ({kai.wang, xj.peng}@siat.ac.cn)}}
\author[3]{Jianfei Yang}
\author[3]{Shijian Lu}
\author[1]{Yu Qiao\thanks{Corresponding author (yu.qiao@siat.ac.cn)}}
\affil[1]{ShenZhen Key Lab of Computer Vision and Pattern Recognition, SIAT-SenseTime Joint Lab,Shenzhen Institutes of Advanced Technology, Chinese Academy of Science}
\affil[2]{University of Chinese Academy of Sciences, China}
\affil[3]{ Nanyang Technological University Singapore}

%	{\tt\small kai.wang@siat.ac.cn}
	%\and
	%Xiaojiang Peng\\
	%Shenzhen Institutes of Advanced Technology\\
	%{\tt\small xj.peng@siat.ac.cn}
	%\and
	%Jianfei Yang\\
	%Nanyang Technological University, Singapore\\
	%{\tt\small yang0478@ntu.edu.sg}
	%\and
	%Shijian Lu\\
	%Nanyang Technological University, Singapore\\
	%{\tt\small Shijian.Lu@ntu.edu.sg}
	%\and
	%Yu Qiao\\
	%Shenzhen Institutes of Advanced Technology\\
	%{\tt\small yu.qiao@siat.ac.cn}

\maketitle
\thispagestyle{empty}
%%%%%%%%% ABSTRACT facilitate

\begin{abstract}
	Annotating a qualitative large-scale facial expression dataset is extremely difficult due to the uncertainties caused by ambiguous facial expressions, low-quality facial images, and the subjectiveness of annotators. These uncertainties lead to a key challenge of large-scale Facial Expression Recognition (FER) in deep learning era. To address this problem, this paper proposes a simple yet efficient Self-Cure Network (SCN) which suppresses the uncertainties efficiently and prevents deep networks from over-fitting uncertain facial images.
	Specifically, SCN suppresses the uncertainty from two different aspects: 1) a self-attention mechanism over mini-batch to weight each training sample with a ranking regularization, and 2) a careful relabeling mechanism to modify the labels of these samples in the lowest-ranked group. Experiments on synthetic FER datasets and our collected WebEmotion dataset validate the effectiveness of our method. 
	Results on public benchmarks demonstrate that our SCN outperforms current state-of-the-art methods with \textbf{88.14}\% on RAF-DB, \textbf{60.23}\% on AffectNet, and \textbf{89.35}\% on FERPlus. The code will be available at \href{https://github.com/kaiwang960112/Self-Cure-Network}{https://github.com/kaiwang960112/Self-Cure-Network}.

\end{abstract}

 \section{Introduction}
 \label{intro}
 \input{introduction}

 \section{Related Work}
 \input{relatedwork}

 \section{Self-Cure Network}
 \label{method}
 \input{RankCleaner}

 \section{Experiments}
 \label{method}
 \input{Experiments}

 \section{Conclusion}
 \label{conclusion}
 \input{Conclusion}
 
 \section{Acknowledge}
 {This work is partially supported byScience and Technology Service Network Initiative of Chinese Academy of Sciences (KFJ-STS-QYZX-092), Guangdong Special Support Program (2016TX03X276), and National Natural Science Foundation of China (U1813218, U1713208), Shenzhen Basic Research Program (JCYJ20170818164704758, CXB201104220032A), the Joint Lab of CAS-HK.}

%-------------------------------------------------------------------------
{\small
	\bibliographystyle{ieee_fullname}
	\bibliography{reference}
}

\end{document}

%% file: introduction.tex
Facial expression is one of the most natural, powerful and universal signals for human beings to convey their emotional states and intentions \cite{darwin1998expression,tian2001recognizing}. Automatically recognizing facial expression is also important to help the computer understand human behavior and to interact with them. In \pxj{the past decades,  researchers have made significant progress on facial expression recognition (FER) with algorithms and large-scale datasets, where datasets can be collected in laboratory or in the wild}, such as CK+ \cite{lucey2010extended}, MMI \cite{valstar2010induced}, Oulu-CASIA \cite{zhao2011facial}, SFEW/AFEW \cite{dhall2011static}, FERPlus \cite{BarsoumICMI2016}, AffectNet \cite{mollahosseini2017affectnet}, EmotioNet \cite{fabian2016emotionet}, RAF-DB \cite{li2017reliable}, etc. %Masses of FER approaches are proposed and achieve the state-of-the-art performance base on these datasets.
\begin{figure}[tp]
    \setlength{\abovecaptionskip}{0.cm}
	\includegraphics[width=0.48\textwidth]{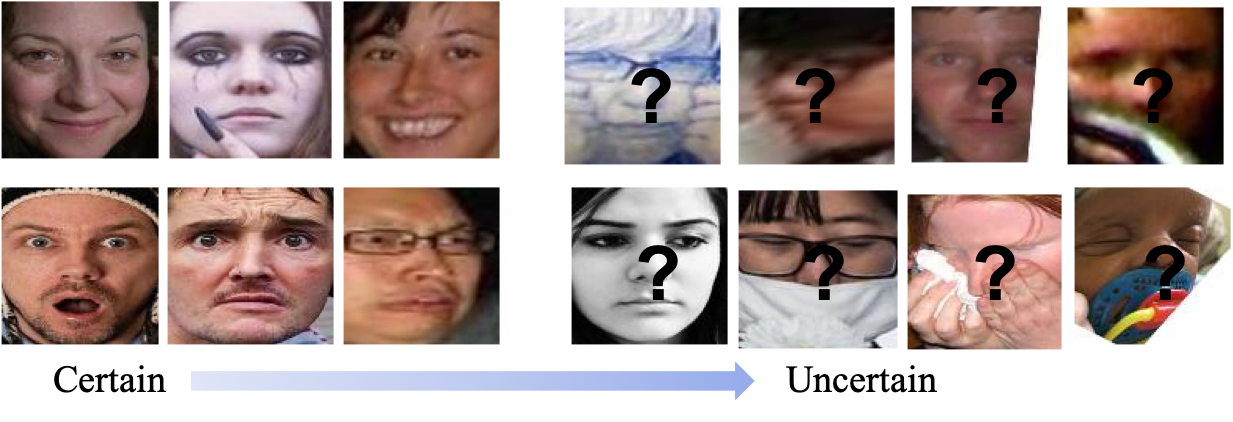}
	\caption{Illustration of uncertainties on real-world facial images from RAF-DB. The right samples are extremely difficult for machines and even human which are better to be suppressed in training.}
	\label{fig:moti}
\end{figure}

However, for the large-scale FER datasets collected from the Internet, it is extremely difficult to annotate with high quality due to the uncertainties caused by the subjectiveness of annotators as well as ambiguous in-the-wild facial images. As illustrated in Figure \ref{fig:moti}, the uncertainties increase from high-quality and evident facial expressions to low-quality and micro expressions. These uncertainties usually lead to inconsistent labels and incorrect labels, which are suspending the progress of large-scale Facial Expression Recognition (FER), especially for the one of data-driven deep learning based FER.
Generally, training with uncertainties of FER may lead to the following problems. First, it may result in over-fitting on the uncertain samples which may be mislabeled. Second, it is harmful for a model to learn useful facial expression features. Third, a high ratio of incorrect labels even makes the model disconvergence in the early stage of optimization.

To address these issues, we propose a simple yet efficient method, termed as Self-Cure Network (SCN), to suppress the uncertainties for large-scale facial expression recognition. The SCN consists of three crucial modules: self-attention importance weighting, ranking regularization, and noise relabeling. Given a batch of images, a backbone CNN is first used to extract facial features. Then the self-attention importance weighting module learns a weight for each image to capture the sample importance for loss weighting. It is expected that uncertain facial images are assigned low importance weights. Further, the ranking regularization module ranks these weights in descending order, splits them into two groups (i.e. high importance weights and low importance weights), and regularizes the two groups by enforcing a margin between the average weights of the two groups. This regularization is implemented with a loss function, termed as Rank Regularization loss (RR-Loss). The ranking regularization module ensures that the first module learns meaningful weights to highlight certain samples (\eg reliable annotations) and to suppress uncertain samples (\eg ambiguous annotations).
The last module is a careful relabeling module that attempts to relabel these samples from the bottom group by comparing the maximum predicted probabilities to the probabilities of given labels. A sample is assigned to a pseudo label if the maximum prediction probability is higher than the one of given label with a margin threshold.
In addition, since the main evidence of uncertainties is the incorrect/noisy annotation problem, we collect an extreme noisy FER dataset from the Internet, termed as WebEmotion, to investigate the effect of SCN with extreme uncertainties. 

Overall, our contributions can be summarized as follows,
\begin{itemize}
\item We innovatively pose the uncertainty problem in facial expression recognition, and propose a Self-Cure Network to reduce the impact of uncertainties.
\item We elaborately design a rank regularization to supervise the SCN to learn meaningful importance weights, which also provides a reference for the relabeling module.
\item We extensively validate our SCN on synthetic FER data and a new real-world uncertain emotion dataset (WebEmotion) collected from the Internet. Our SCN also achieves performance \textbf{88.14}\% on RAF-DB, \textbf{60.23}\% on AffectNet, and \textbf{89.35}\% on FERPlus, which set new records on them.
\end{itemize}
%on public benchmarks demonstrate that our SCN outperforms current state-of-the-art methods with \textbf{88.14}\% on RAF-DB, \textbf{60.23}\% on AffectNet, and \textbf{89.35}\% on FERPlus.
%\begin{itemize}
%\item To We propose Self-Cure Network (SCN) to reduce the influences of noisy annotations and relabel the noisy samples simultaneously. 
%%SCN includes three new modules, namely self-attention importance weighting, ranking regularization, and noise relabeling. \pxj{These plug-and-play  modules effectively address network training under noisy annotations.}
%
%\item We elaborately design a rank regularization to ensure the SCN to learn robust features with noisy annotations more efficiently, \pxj{which also provides a reference for noise relabeling module (i.e. relabeling samples in low-importance group).}
%
%\item \pxj{ In addition to synthetic FER data, we also collect a real-world noisy emotion dataset (WebEmotion) from the Internet to validate our proposed SCN. We conduct extensive experiments on RAF-DB, FERPlus, and AffectNet, and obtain promising improvements.} Our method  achieves state-of-the-art results on these datasets.
%\end{itemize}

%% file: relatedwork.tex
\subsection{Facial Expression Recognition}
Generally, a FER system mainly consists of three stages, namely face detection, feature extraction, and expression recognition. 
In face detection stage, several face detectors like MTCNN~\cite{7553523} and Dlib~\cite{amos2016openface}) are used to locate faces in complex scenes. The detected faces can be further aligned alternatively. For feature extraction, various methods are designed to capture facial geometry and appearance features caused by facial expressions. According to the feature type, they can be grouped into engineered features and learning-based features. For the engineered features, they can be further divided into texture-based local features, geometry-based global features, and hybrid features. The texture-based features mainly include SIFT~\cite{sift}, HOG~\cite{1467360}, Histograms of LBP~\cite{SHAN2009803}, Gabor wavelet coefficients~\cite{999679}, etc. The geometry-based global features are mainly based on the landmark points around noses, eyes, and mouths. Combining two or more of the engineered features refers to the hybrid feature extraction, which can further enrich the representation. 
For the learned features, Fasel~\cite{Fasel2002Robust} finds that a shallow CNN is robust to face poses and scales. Tang~\cite{Tang2013Deep} and Kahou \textit{et al}.~\cite{Kahou2013Combining} utilize deep CNNs for feature extraction, and win the FER2013 and Emotiw2013 challenge, respectively. Liu \textit{et al}.~\cite{Liu2015AU} propose a Facial Action Units based CNN architecture for expression recognition. 
Recently, both Li \textit{et al.}~\cite{8576656} and Wang \textit{et al.}~\cite{wang2019region} have designed region-based attention networks for pose and occlusion aware FER, where the regions are either cropped from landmark points or fixed positions.

\subsection{Learning with Uncertainties}
Uncertainties in the FER task mainly come from ambiguous facial expressions, low-quality facial images, inconsistent annotations, and incorrect annotations (\ie noisy labels). Particularly, learning with noisy labels is extensively studied in the computer vision community while the other two aspects are rarely explored. In order to handle noisy labels, one intuitive idea is to leverage a small set of clean data that can be used to assess the quality of the labels during the training process \cite{veit2017learning,li2017learning,dehghani2017avoiding}, or to estimate the noise distribution \cite{sukhbaatar2014learning}, or to train the feature extractors \cite{azadi2015auxiliary}. Li \textit{et al}. \cite{li2017learning} propose a unified distillation framework using `side' information from a small clean dataset and label relations in knowledge graph, to `hedge the risk' of learning from noisy labels. Veit \textit{et al}.\cite{Veit_2017_CVPR} use a multi-task network that jointly learns to clean noisy annotations and to classify images. Azadi \textit{et al}.\cite{azadi2015auxiliary} select reliable images by an auxiliary image regularization for deep CNNs with noisy labels. Other methods do not need a small clean dataset but they may assume extra constrains or distributions on the noisy samples \cite{mnih2012learning}, such as a specific loss for randomly flipped labels \cite{natarajan2013learning}, regularizing the deep networks on corrupted labels by a MentorNet \cite{jiang2017mentornet}, and other approaches that model the noise with a softmax layer by connecting the latent correct labels to the noisy ones \cite{goldberger2016training,zeng2018facial}. 
For the FER task, Zeng \textit{et al}.~\cite{zeng2018facial} first consider the inconsistent annotation problem among different FER datasets, and propose to leverage these uncertainties to improve FER. \textit{In contrast, our work focuses on suppressing these uncertainties to learn better facial expression features.}

%% file: RankCleaner.tex
\begin{figure*}[htp]
    \setlength{\abovecaptionskip}{0.cm}
	\includegraphics[width=0.90\textwidth]{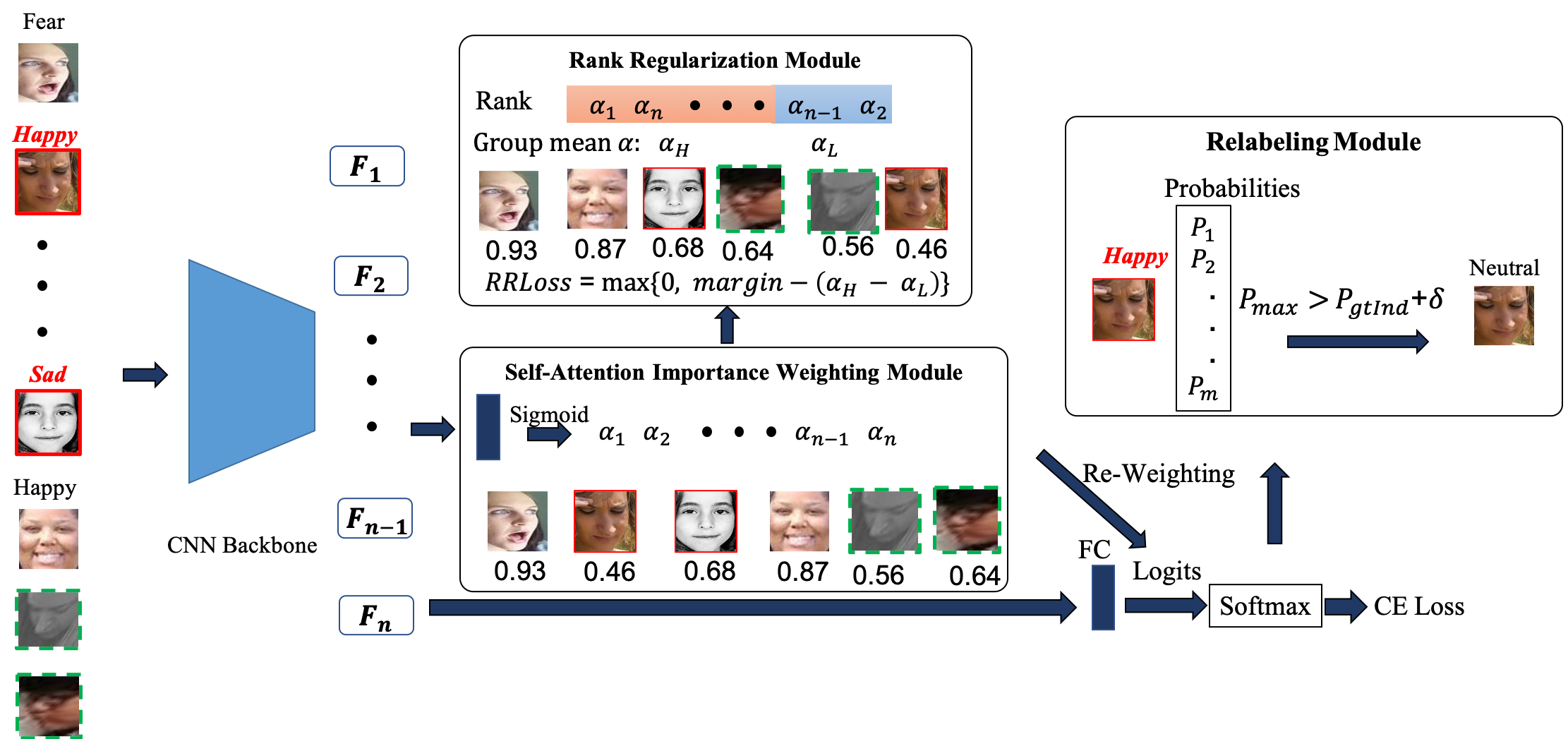}
	\caption{The pipeline of our Self-Cure Network. Face images are first fed into a backbone CNN for feature extraction. The self-attention importance weighting module learns sample weights from facial features for loss weighting. The rank regularization module takes as input the sample weights and constrain them with a ranking operation and a margin-based loss function. The relabeling module hunts reliable samples by comparing maximum predicted probabilities to the probabilities of given labels. \textit{Mislabeled samples are marked in red solid rectangles and ambiguous samples in green dash ones. It is worth noting that SCN mainly resorts to the re-weighting operation to suppress these uncertainties and only modifies some of the uncertain samples}. }
	\label{fig:pipeline}
	\setlength{\belowcaptionskip}{0.cm}
\end{figure*}

\pxj{To learn robust facial expression features with uncertainties, we propose a simple yet efficient Self-Cure Network (SCN). In this section, we first provide an overview of the SCN, and then present its three modules. We finally present the detailed implementation of SCN.}

\subsection{Overview of Self-Cure Network}

%The annotation noises can affect the performance and the feature learning of CNNs in many deep learning tasks including the facial expression recognition. Most of the facial expression datasets suffer from the annotation noises, which results in a low recognition performance (accuracy). The proposed Self-Cure Network (SCN) is designed to overcome this problem. 
Our SCN is built upon traditional CNNs and consists of three crucial modules: i) self-attention importance weighting, ii) ranking regularization, and iii) relabeling, as shown in Figure \ref{fig:pipeline}.  

Given a batch of face images with some uncertain samples, we first extract the deep features by a backbone network. The self-attention importance weighting module assigns an importance weight for each image using a fully-connected (FC) layer and the sigmoid function. These weights are multiplied by the logits for a sample re-weighting scheme. To explicitly reduce the importance of uncertain samples, a rank regularization module is further introduced to regularize the attention weights. In the rank regularization module, we first rank the learned attention weights and then split them into two groups, i.e. high and low importance groups. We then add a constraint between the mean weights of these groups by a margin-based loss, which is called rank regularization loss (RR-Loss). 
 To further improve our SCN, the relabeling module is added to modify some of the uncertain samples in the low importance group. This relabeling operation aims to hunt more clean samples and then to enhance the final model. The whole SCN can be trained in an end-to-end manner and easily added into any CNN backbones.
%The weighted Cross-Entropy loss is used to reduce the influence of noisy annotations by weighting on the logits. To further improve our SCN performance, noise relabeling aims to correct the samples from the low importance group. The whole SCN can be trained in an end-to-end manner and easily added into many CNNs backbones.

\subsection{Self-Attention Importance Weighting}
We introduce the self-attention importance weighting module to capture the contributions of samples for training. It is expected that certain samples may have high importance weights while uncertain ones have low importance. 
Let $\mathbf{F}=[\mathbf{x}_1, \mathbf{x}_2, \ldots, \mathbf{x}_N]\in R^{D\times N}$ denotes the facial features of $N$ images, the self-attention importance weighting module takes $\mathbf{F}$ as input, and  output an importance weight for each feature. Specifically, the self-attention importance weighting module is comprised of a linear fully-connected (FC) layer and a sigmoid activation function, which can be formulated as ,

\begin{equation}
\alpha_i =\sigma( \mathbf{W}_a^\top \mathbf{x}_i ),
\end{equation}
where $\alpha_i$ is the importance weight of the \textit{i}-th sample, $\mathbf{W}_a$ is the parameters of the FC layer used for attention, and $\sigma$ is the sigmoid function. This module also provides reference for the other two modules.

%Specifically, the extracted features are fed into a fully-connected (FC) layer and a sigmoid function. The size of the feature matrix is $N \times D$, $N$ is the batch size, $D$ is the dimension of features. The input and output channel of the FC layer are $D$ and 1, respectively. To model the importance of each image, we choose sigmoid function as the active function to obtain a value ranging from 0 to 1, the value of which represents the importance of each face. Noise and hard samples are expected to have low importance value. In this module, we expect to learn more diversity of the importance weights among the whole batch images. The weights are also regarded as a critical reference for another two modules. Self-Attention Importance Weighting Module can be formulated as:
%
%\begin{equation}
%\alpha = [\alpha_1, \alpha_2, ... , \alpha_n] = [s(F_1; \theta), s(F_2; \theta), ... , s(F_n; \theta)],
%\end{equation}
%
%where $\alpha_i$ and $F_i$ are the importance weight and feature of each image, $s(\cdot; \theta)$ is the operation of self-attention importance weighting, $\theta$ is the parameter of the $s(\cdot; \theta)$.

\textbf{Logit-Weighted Cross-Entropy Loss}.
With the attention weights, we have two simple choices to perform loss weighting inspired by \cite{hu2019noise}. The first choice is to multiply the weight of each sample by the sample loss. In our case, since the weights are optimized in an end-to-end manner and are learned from the CNN features, they are doomed to be zeros as this trival solution makes zero loss. MentorNet \cite{jiang2017mentornet} and other self-paced learning methods~\cite{jiang2014self,ma2017self} solve this problem by alternating minimization, i.e. optimize one at a time while the other is held fixed. In this paper, we choose the logit-weighted one of \cite{hu2019noise} which is shown to be more efficient. 
For a multi-class Cross-Entropy loss, we call our weighted loss as Logit-Weighted Cross-Entropy loss (WCE-Loss), which is formulated as, 

\begin{equation}
    \mathcal{L}_{WCE} = -\frac{1}{N}\sum_{i=1}^{N}log\frac{e^{\alpha_{i}\mathbf{W}_{y_i}^\top \mathbf{x}_i }}{\sum_{j=1}^{C}e^{^{\alpha_{i}\mathbf{W}_{j}^\top \mathbf{x}_i }}},
\end{equation}
where $\mathbf{W}_j$ is the \textit{j}-th classifier. As suggested in \cite{liu2017sphereface}, the $\mathcal{L}_{WCE}$ has a positive correlation with the $\alpha$. % it can reduce the influence of noisy labels with small weights. %to CNNs since the noisy labels' loss is lower than the correct labels' loss.

\subsection{Rank Regularization}
The self-attention weights in the above module can be arbitrary in (0, 1). To explicitly constrain the importance of uncertain samples, we elaborately design a rank regularization module to regularize the attention weights. In the rank regularization module, we first rank the learned attention weights in descending order and then split them into two groups with a ratio $\beta$. The rank regularization ensures that the mean attention weight of high-importance group is higher than the one of low-importance group with a margin. Formally, we define a rank regularization loss (RR-Loss) for this purpose as follows, 
\begin{equation}
\mathcal{L}_{RR} = max\{0, \delta_1 - (\alpha_H- \alpha_L)\},
\end{equation}
with
\begin{equation}
\alpha_H = \frac{1}{M}\sum_{i=0}^{M}\alpha_{i}, \alpha_L = \frac{1}{N-M}\sum_{i=M}^{N}\alpha_{i},
\end{equation}
where $\delta_1$ is a margin which can be a fixed hyper parameter or a learnable parameter, $\alpha_H$ and $\alpha_L$ are the mean values of the high importance group with $\beta * N = M$ samples and the low importance group with $N-M$ samples, respectively. 
In training, the total loss function is $\mathcal{L}_{all} = \gamma \mathcal{L}_{RR} + (1-\gamma) \mathcal{L}_{WCE}$ where $\gamma$ is a trade-off ratio. %  RR-Loss is jointly optimized with the WCE-Loss. %These two losses prevent $\alpha$ from falling into trivial solution and reduce the importance weights of noisy samples. %Furthermore, the WCE-Loss can be also added into other classification tasks. 

%
%We then add a constraint between the mean weights of these groups by a margin-based loss, which is called rank regularization loss (RR-Loss). 
%To prevent the weights from degrading to the same value, we should impose a regularization on it. Imaging all the weights with the same value, we can not discriminate the importance divergence by the weights. Besides, the divergence of the weights can help us mining the noisy annotations.
%%Although we design a module to learn the importance weights of images, the weights may fall into a same value without any other constraint. For example, all of the importance weights are 0.5, which can not show the difference between the batch images. 
%Therefore, we design a Rank Regularization Loss (RR-Loss) as a constraint to highlight the images with high importance values and to reduce the influence of the images with low importance values. We divide the features of a batch images into two groups, one group for high importance samples and the other for low importance samples. Consequently, RR-Loss can ensure a margin between two groups. The RR-Loss is formally defined as,
%
%\begin{equation}
%L_{RR} = max\{0, \delta - (H_{\alpha}- L_{\alpha})\},
%\end{equation}
%\begin{equation}
%H_{\alpha} = \frac{\sum_{i=0}^{N-M}\alpha_{i}}{N-M}, L_{\alpha} = \frac{\sum_{i=N-M}^{N}\alpha_{i}}{M},
%\end{equation}
%
%where $\delta$ is a margin between the two groups, $H_{\alpha}$ and $L_{\alpha}$ are the mean value of the high importance group ($N-M$ samples) and low importance groups ($M$ samples). 

\begin{table*}[t]
\center
\setlength{\abovecaptionskip}{0.cm}
\caption{The statistics of our WebEmotion.}
%\resizebox{\linewidth}{!}{
\begin{tabular}{cccccccccccc}
\toprule
Category & Happy & Sad & Surprise & Fear & Angry & Disgust & Contempt & Neutral & \textbf{Total} \\ \midrule
\# Videos & 4,231 & 5,670 & 4,573 & 5,328 & 5,668 & 5,197 & 5,266 & 5,406 & \textbf{41,339}\\
\# Clips  & 27,854 & 29,667 & 27,418 & 29,822 & 31,483 & 20,764 & 6,454 & 26,687 & \textbf{200,149}\\ 
\bottomrule
\end{tabular}%}
\label{tab:dataset}
\end{table*}

\subsection{Relabeling}
In the rank regularization module, each mini-batch is divided into two groups, i.e. the high-importance and the low-importance groups. We experimentally find that the uncertain samples usually have low importance weights, thus an intuitive idea is to design a strategy to relabel these samples. The main challenge to modify these annotations is to know which annotation is incorrect. 

Specifically, our relabeling module only considers the samples in the low-importance group and is performed on the Softmax probabilities. For each sample, we compare the maximum predicted probability to the probability of given label. A sample is assigned to a new pseudo label if the maximum prediction probability is higher than the one of given label with a threshold. Formally, the relabeling module can be defined as,
\begin{equation}
y' = \left\{ \begin{array}{ll}
l_{max} & \textrm{if $P_{max} - P_{gtInd} > \delta_2$},\\
l_{org} & \textrm{ otherwise},\\
\end{array} \right.
\end{equation}
where $y'$ denotes the new label, $\delta_2$ is a threshold, $P_{max}$ is the maximum predicted probability, and $P_{gtInd}$ is the predicted probability of the given label. $l_{org}$ and $l_{max}$ are the original given label and the index of the maximum prediction, respectively.

In our system, uncertain samples are expected to obtain low importance weights thus to degrade their negative impacts with re-weighting, and then fall into the low-importance group, and finally may be corrected as certain samples by relabeling. Those corrected samples may obtain high important weights in the next epoch. \textit{We expect the network can be cured by itself with either re-weighting or relabeling, which is the reason why we call our method as self-cured network.}
%If $S_{max}$ - $S_{org}$ $>$ $\phi$ (default: 0.2), we will change the label. If we correct a noise sample, the sample may obtain a high importance weight in the next epoch. We expect the network can be cured itself during the training stage.

\subsection{Implementation}
\textbf{Pre-processing and facial features}. In our SCN, face images are detected and aligned by MTCNN \cite{zhang2016joint} and further resized to 224 $\times$ 224 pixels. The SCN is implemented with Pytorch toolbox and the backbone network is ResNet-18 \cite{he2016deep}. By default, the ResNet-18 is pre-trained on the MS-Celeb-1M face recognition dataset and the facial features are extracted from its last pooling layer. 

\textbf{Training}. We train our SCN  in an end-to-end manner with 8 Nvidia Titan 2080ti GPU, and set the batch size as 1024. In each iteration, the training images are divided into two groups including 70\% high importance samples and 30\% low importance samples by default. The margin $\delta_1$ between the mean value of high and low importance groups can be either set at 0.15 by default or designed as a learnable parameter. Both strategies will be evaluated in the ensuing Experiments. The whole network is jointly optimized with RR-Loss and WCE-Loss. The ratio of the two losses is empirically set at 1:1, and its influence will be studied in the ensuing ablation study of Experiments. The leaning rate is initialized as 0.1 which is further divided by 10 after 15 epochs and 30 epochs, respectively. The training stops at 40 epochs.
The relabeling module is included for optimization from the 10th epoch, where the relabeling margin $\delta_2$ is set at 0.2 by default. 

%% file: Experiments.tex
In this section, we first describe three public datasets and our WebEmotion dataset. We then demonstrate the robustness of our SCN under uncertainties of both synthetic and real-world noisy facial expression annotations. Further, we conduct ablation studies with qualitative and quantitative results to show the effectiveness of each module in SCN. Finally, we compare our SCN to the state-of-the-art methods on public datasets. %urther explore the effect of each component of SCN and compare our method with the prevailing approaches.

%This section  presents experimental results including the datasets evaluated, the qualitative and quantitative experimental results as well as ablation studies. %we employed and our own Web-Emotion dataset specially collected for large-scale facial expression recognition with noises.

\subsection{Datasets}
%We evaluate our SCN over three popular in-the-wild FER datasets, namely RAF-DB, AffectNet, and FERPlus. Beyond that, twe also evaluate over a Web-Emotion dataset that we collected for large-scale FER with annotation noises.

\textbf{RAF-DB}~\cite{li2017reliable} contains 30,000 facial images annotated with basic or compound expressions by 40 trained human coders. In our experiment, only images with six basic expressions (neutral, happiness, surprise, sadness, anger, disgust, fear) and neutral expression are used which leads to 12,271 images for training and 3,068 images for testing. The overall sample accuracy is used for measurement.

\textbf{FERPlus}~\cite{BarsoumICMI2016} is extended from FER2013 as used in the \textit{ICML 2013 Challenges}. It is a large-scale dataset collected by the Google search engine. It consists of 28,709 training images, 3,589 validation images and 3,589 test images, all of which are resized to 48$\times$48 pixels. \textit{Contempt} is included which leads to 8 classes in this dataset. The overall sample accuracy is used for measurement %FERPlus differs from FER2013 mainly by annotation, where FER2013 has the same seven expression labels as RAF-DB while FERPlus has one more \textit{contempt} label.

\textbf{AffectNet}~\cite{mollahosseini2017affectnet} is by far the largest dataset that provides both categorical and Valence-Arousal annotations. It contains more than one million images from the Internet by querying expression-related keywords in three search engines, of which 450,000 images are manually annotated with eight expression labels as in FERPlus. It has imbalanced training and test sets as well as a balanced validation set. The \textit{mean class accuracy} on the validation set is used for measurement.

\textbf{The collected WebEmotion}.
Since the main evidence of uncertainties is the incorrect/noisy annotation problem, we collect an extreme noisy FER dataset from the Internet, termed as WebEmotion, to investigate the effect of SCN with extreme uncertainties.
The WebEmotion is a video dataset (though we use it as image data by assigning labels to frames) downloaded from YouTube with a set of keywords including 40 emotion-related words, 45 countries from \textit{Asia, Europe, Africa, America}, and 6 age-related words (i.e. \textit{baby, lady, woman, man, old man, old woman}). It consists of the same 8 classes with FERPlus, where each class is connected to several emotion-related keywords, \eg \textit{Happy} is connected to the keywords \textit{happy, funny, ecstatic, smug, and kawaii}. To obtain meaningful correlation between the keywords and the searched videos, only the top 20 crawled videos with less then 4 minutes are selected. This leads to around 41,000 videos which are further segmented into 200,000 video clips with a constraint that a face (detected by MTCNN) appears at least 5 seconds. %Finally, face images are extracted from the collected video clips by using face detector \cite{zhang2016joint} and labelled according to the emotion keywords. 
For evaluation, we only use WebEmotion for pretraining since annotating is extremely difficult.
Table \ref{tab:dataset} shows the statistics of WebEmotion. The meta videos and video clips will be public to the research community.% for uncertain FER.

% What is MS-1M?
% The baseline is not properly defined with sufficient details
% $\checkmark \checkmark$ is pre-trained over Web-Emotion and then fine-tuned by Web-Emotion? - Shijian

\begin{figure*}[h]
\setlength{\abovecaptionskip}{0.cm}
\center
	\includegraphics[width=0.8\textwidth]{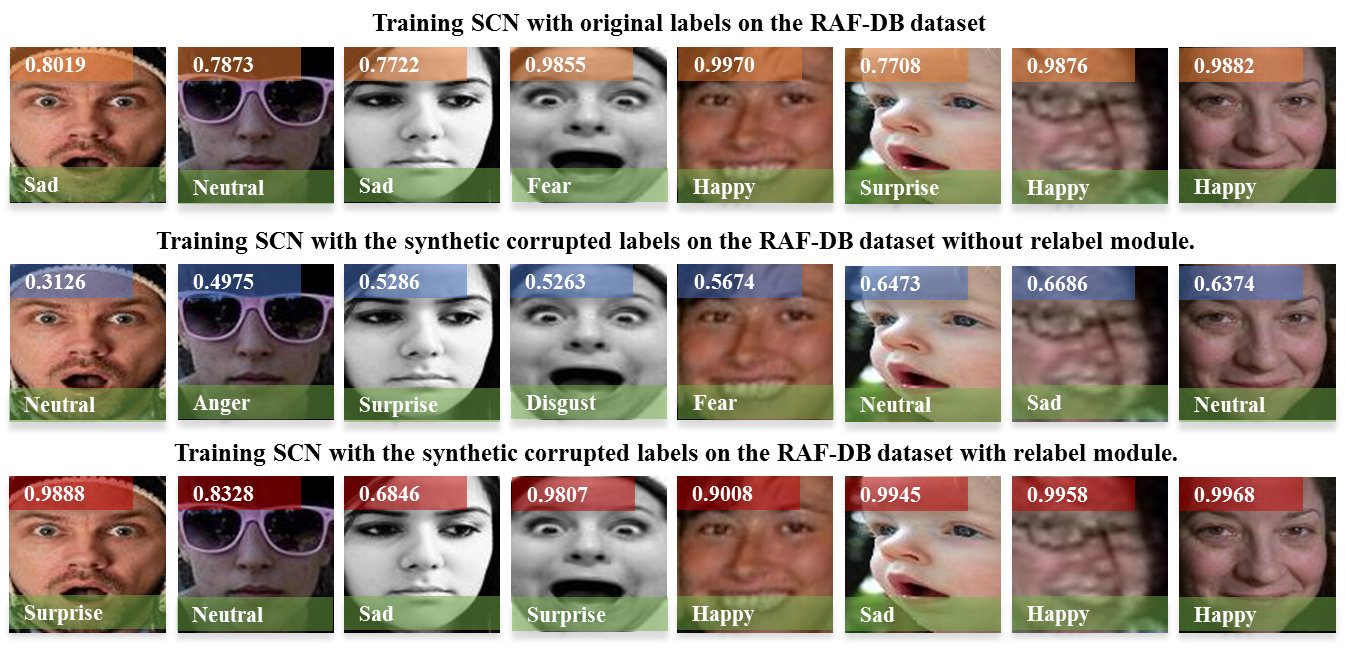}
	\caption{Visualization of the learned importance weights in our SCN, we show these weights on \pxj{randomly selected images with original labels (1st row) and synthetic noisy labels before and after relabeling (2nd row and 3rd row).}}
	\label{fig:alpha}
\end{figure*}

\subsection{Evaluation of SCN on Synthetic Uncertainties}
%\pxj{Table \ref{tab:WITH_NOISE}}
The uncertainties of FER mainly come from ambiguous facial expressions, low-quality facial images, inconsistent annotations, and incorrect annotations (\ie noisy labels). Considering that only noisy labels can be analyzed quantitatively, we explore the robustness of SCN with three levels of label noises including the ratio of 10\%, 20\%, and 30\% to RAF-DB, FERPLus, and AffectNet datasets. \pxj{Specifically, we randomly choose 10\%, 20\%, and 30\% of training data for each category and randomly change their labels to others.} In Table \ref{tab:WITH_NOISE}, we use ResNet-18 as CNN backbone and compare our SCN to the baseline (traditional CNN training without considering label noises) with \pxj{two training schemes: i) training from scratch and ii) fine-tuning with a pretrained model on Ms-Celeb-1M~\cite{GuoZHHG16}. } We also compare our SCN to two state-of-the-art noise-tolerant methods on RAF-DB, namely CurriculumNet~\cite{Guo_2018_ECCV} and MetaCleaner~\cite{Zhang_2019_CVPR}. 

\pxj{As shown in Table \ref{tab:WITH_NOISE}, our SCN consistently improves the baseline by a large margin. 
For scheme i) with noise ratio 30\%, our SCN outperforms the baseline by 13.80\% , 1.07\%, and 1.91\% on RAF-DB, FERPLus, and AffectNet, respectively. 
For scheme ii) with noise ratio 30\%, our SCN still gain improvements of 2.20\%, 2.47\%, and 3.12\% on these datasets though the performance of them are relatively high.  For both schemes, the benefit from SCN becomes more obvious as the noise ratio increases up.
CurriculumNet designs training curriculum by measuring data complexity using cluster density which can avoid training noisy-labeled data in early stages. MetaCleaner aggregates the features of several samples in each class into a weighted mean feature for classification which can also weaken the noisy-labeled samples. Both CurriculumNet and MetaCleaner improve the baseline largely but are still inferior to the SCN which is simpler.
Another interesting find is that the improvement of SCN on RAF-DB is much higher than on other datasets. 
It may be explained by the following reasons. On the one hand, RAF-DB consists of compound facial expressions and is annotated by 40 people with crowdsourcing, which make the data annotations more inconsistent. Thus, our SCN may also gain improvement on the original RAF-DB without synthetic label noises. On the other hand, AffectNet and FERPlus are annotated by experts, thus less inconsistent labels are involved, leading to less improvement on RAF-DB.}
%These results illustrate our SCN has strong noise-tolerant ability on these synthetic noises datasets, and it achieves improvement on both direct training or finetuning settings.  The proof is the results of RAF-DB dataset, the reason of which may come from two perspectives. On the one hand, RAF-DB datasets is annotated by 40 people by crowdsourcing, which may lead many annotation inconsistently. One the other hand, the RAF-DB dataset consists of compound facial expressions, which increases the FER difficulties. As the AffectNet and FERPLus are annotated by the experts, less noisy labels are involved, leading to less improvement than that of the RAF-DB dataset.
% As for the AffectNet validation that includes 5000 labeled images in 10 categories, we evaluate baseline method and SCN with 3500 images in 8 categories. FERPLus is a dataset with high consistently, because 10 people were trained for better annotate FERPLus. Because of the different annotation and evaluation methods, our SCN gains different improvement.

\begin{table}[t!]
\center
\setlength{\abovecaptionskip}{0.cm}
\caption{The evaluation of SCN on synthetic noisy FER datasets. `Pretrain' means we use a pretrained model from face recognition, otherwise we train from scratch.}
\resizebox{\linewidth}{!}{
\begin{tabular}{@{}ccccccccc@{}}
\toprule
 Pretrain & SCN  & Noise(\%) & RAF-DB  & AffectNet & FERPlus\\ \midrule
$\times$  &  CurriculumNet~\cite{Guo_2018_ECCV} & 10 & 68.5 &- &-\\ 
$\times$  &  MetaCleaner~\cite{Zhang_2019_CVPR} & 10 & 68.45 &- &-\\ 
 $\times$     & $\times$  & 10 & 61.43 & 44.68 & 77.15 \\
$\times$  & $\checkmark$  & 10 &  \textbf{70.26} & \textbf{45.23} &\textbf{78.53} \\ \midrule
 $\times$  &  CurriculumNet~\cite{Guo_2018_ECCV} & 20 & 61.23 &- &-\\ 
  $\times$  &  MetaCleaner~\cite{Zhang_2019_CVPR} & 20 & 61.35 &- &-\\ 
 $\times$     & $\times$ & 20 & 55.5 &  41.00 & 71.88 \\
 $\times$  & $\checkmark$   & 20 & \textbf{63.50} &\textbf{41.63} &\textbf{72.46} \\ \midrule
 $\times$  &  CurriculumNet~\cite{Guo_2018_ECCV} & 30 & 57.52 & - &-\\ 
  $\times$  &  MetaCleaner~\cite{Zhang_2019_CVPR} & 30 & 58,89 & - &-\\ 
 $\times$     & $\times$  & 30   & 46.81 &  38.35 & 68.54 \\
 $\times$  & $\checkmark$   & 30 & \textbf{60.61} & \textbf{39.42} & \textbf{70.45} \\
 \midrule
$\checkmark$ & $\times$  & 10 & 80.81 & 57.18 & 83.39 \\
$\checkmark$ & $\checkmark$  & 10 & \textbf{82.18} & \textbf{58.58} & \textbf{84.28} \\ 
$\checkmark$ & $\times$   & 20 & 78.18 & 56.15 & 82.24 \\
 $\checkmark$ & $\checkmark$   & 20 & \textbf{80.10} & \textbf{57.25} & \textbf{83.17} \\ 
 $\checkmark$ & $\times$   & 30  & 75.26 & 52.58 & 79.34 \\
$\checkmark$ & $\checkmark$   & 30 & \textbf{77.46} & \textbf{55.05} & \textbf{82.47} \\
\bottomrule
\end{tabular}}
\label{tab:WITH_NOISE}
\end{table}

\textbf{Visualization of $\alpha$ in SCN}. 
\pxj{To further investigate the effectiveness of our SCN under noisy annotations, we visualize the importance weight $\alpha$ during the training phase of SCN on RAF-DB with noise ratio 10\% . In Figure \ref{fig:alpha}, the first row indicates the importance weights when SCN is trained with original labels. The images of the second row are annotated with synthetic corrupted labels, and we use SCN (without Relabel module) to train the synthetic noisy dataset. Indeed, the SCN regards those label-corrupted images as noises and automatically suppresses the weights of them. After sufficient training epochs, the relabeling module are added into SCN, and these noisy-labeled images are relabeled (of course many others may be not relabeled since we have relabeling constraint). After several other epochs, the importance weights of them become high (the 3rd row), which demonstrates that our SCN can `self-cure' the corrupted labels.} It is worth noting that the new labels from relabeling module may be inconsistent with ``ground-truth'' labels (see the 1st, 4th, and 6th columns) but they are also reasonable in visualization.

\begin{table}[t!]
\center
\setlength{\abovecaptionskip}{0.cm}
\caption{The effect of SCN on WebEmotion for pretraining. The 2nd column indicates finetuning with or without SCN.}
%\resizebox{\linewidth}{!}{
\begin{tabular}{@{}cccccccc@{}}
\toprule
 WebEmoition & SCN  & RAF-DB  & AffectNet & FERPlus\\ \midrule
$\times$     & $\times$  & 72.00 & 46.58 &82.4 \\
w/o SCN  & $\times$   &  78.97 & 56.43 & 84.20 \\
w/o SCN & $\checkmark$   & 80.42& 57.23 & 85.13 \\
SCN & $\checkmark$   & 82.45 & 58.45 & 85.97 \\ 
\bottomrule
\end{tabular}%}
\label{tab:webemotion}
\end{table}

\subsection{Exploring SCN on Real-World Uncertainties}
Synthetic noisy data proves the effectiveness of the `self-curing' ability of SCN. \pxj{In this section, we apply our SCN to real-world FER datasets which can include all types of uncertainties.}

\textbf{SCN on WebEmotion for pretraining}. %\pxj{Table \ref{tab:webemotion}}
Our collected WebEmotion dataset consists of massive noises since the searching keywords are regarded as labels. \pxj{To better validate the effect of SCN on real-world noisy data, we apply our SCN to WebEmotion for pretraining and then finetune the model on target datasets. We show the comparison experiments in Table \ref{tab:webemotion}. 
From the 1st and the 2nd rows, we can see that pretraining on WebEmotion without SCN improves the baseline by 6.97\%, 9.85\%, and 1.80\% on RAF-DB, FERPlus and AffectNet, respectively. Fine-tuning with SCN on target datasets obtains gains ranged from 1\% to 2\%.  Pretraining on WebEmotion with SCN further boosts the performance from 80.42\% to 82.45\% on RAF-DB.} This suggests that SCN learns robust features on WebEmotion which is better for further fine-tuning.
%Since there are massive annotation noises in WebEmotion dataset, we use the SCN in both WebEmotion and target datasets training. As expected, this training strategy gains \textbf{2.03\%}, \textbf{1.22\%}, and \textbf{0.84\%} on three datasets, respectively.

%In this section, we first compare our SCN to baseline method on original RAF-DB. AffectNet, and FERPlus datasets, and then show the robustness of our proposed SCN on the these datasets with different level of manual added noises.
\begin{figure}[tp]
    \setlength{\abovecaptionskip}{0.cm}
	\includegraphics[width=0.48\textwidth]{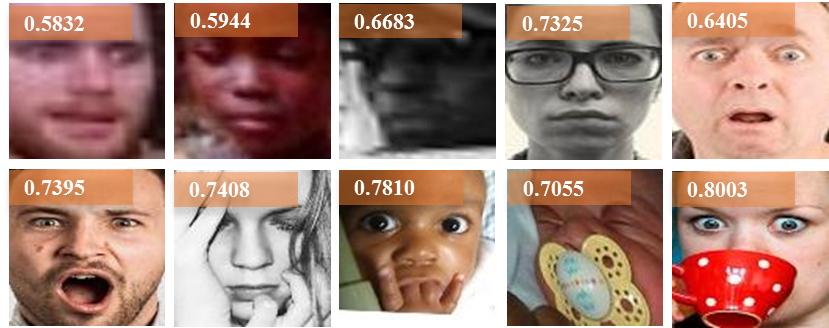}
	\caption{\pxj{Ten examples of RAF-DB (w/o synthetic noisy labels) with low importance weights. Each column corresponds to a basic emotion. One can guess their labels and the ground-truth labels of RAD-DB are included in the text}.}
	\label{fig:samples}
\end{figure}

\begin{table}[t!]
\center
\setlength{\abovecaptionskip}{0.cm}
\caption{SCN on real-world FER datasets. The improvements of SCN suggests that these public datasets more or less suffer from uncertainties. }
\resizebox{\linewidth}{!}{
\begin{tabular}{@{}cccccccc@{}}
\toprule
 Pretrain & SCN   & RAF-DB  & AffectNet & FERPlus\\ \midrule
 $\times$     & $\times$  & 72.00 & 46.58 &82.4 \\
 $\times$  & $\checkmark$   &  \textbf{78.31} & \textbf{47.28} &\textbf{83.42} \\
 $\times$  &  CurriculumNet~\cite{Guo_2018_ECCV} &  74.67 &- &-\\ 
  $\times$  &  MetaCleaner~\cite{Zhang_2019_CVPR} &  77.18 &- &-\\ \midrule
 $\checkmark$ & $\times$   & 84.20 & 58.5 & 86.80 \\
$\checkmark$ & $\checkmark$   & \textbf{87.03} & \textbf{60.23} & \textbf{88.01} \\ 
\bottomrule
\end{tabular}}
\label{tab:ferNoNoise}
\end{table}

\textbf{SCN on Original FER datasets}. %\pxj{Table \ref{tab:ferNoNoise}}
We further conduct experiments on original FER datasets to evaluate our SCN since these datasets inevitably suffer from uncertainties such as  ambiguous facial expressions, low-quality facial images, etc. Results are shown in Table \ref{tab:ferNoNoise}. When training from scratch, our proposed SCN improves the baseline consistently with gains of 6.31\%, 0.7\%, and 1.02\% on RAD-DB, AffectNet, and FERPlus, respectively. MetaCleaner also boosts the baseline on RAF-DB but slightly worse than our SCN.  With pretraining, we still obtain improvements of 2.83\%, 1.73\%, and 1.21\% on these datasets. The improvement of SCN and MetaCleaner suggests that there indeed exists uncertainties in those datasets. To validate our speculation, we rank the importance weights of RAF-DB, and show some examples with low importance weights in Figure \ref{fig:samples}. The ground-truth labels from top-left to bottom-right are \textit{surprise, neutral, neutral, sad, surprise, surprise, neutral, surprise, neutral, surprise}. We find that images with low quality and occlusion are difficult to annotate and are more likely to have low-importance weights in SCN.
%We argue that there are still some confused annotations in RAF-DB though a lot of time and money are used for annotating. 

%It further proves that our SCN can improve recognition performance on real datasets with annotation noises. 
% The noisy annotations or hand samples are always with a low importance weight in the training stage, which means the SCN can suppress the influence of annotation noises.

\begin{table}[t!]
\center
\setlength{\abovecaptionskip}{0.cm}
\caption{Evaluation of the three modules in SCN.}
\resizebox{\linewidth}{!}{
% Evaluation of the three modules in SCN.
\begin{tabular}{@{}cccccc@{}}
\toprule
Weight & Rank & Relabel & RAF-DB & RAF-DB (pretrain) \\ \midrule
 $\times$ & $\times$ & $\times$ & 72.00 & 84.20  \\
 $\times$ & $\times$ & $\checkmark$ & 71.25 & 83.78 \\
 $\times$ & $\checkmark$ & $\times$ & 74.15 & 85.14  \\
 $\checkmark$ & $\times$ & $\times$ & 76.26 & 86.09  \\
 $\checkmark$ & $\checkmark$ & $\times$ & 76.57 & 86.63  \\
$\checkmark$ & $\checkmark$ & $\checkmark$ & 78.31 & 87.03  \\
\bottomrule
\end{tabular}}
\label{tab:modules}
\end{table}

\begin{figure*}[tp]
\setlength{\abovecaptionskip}{0.cm}
\center
%\subfigure[$\delta_1$]{
\includegraphics[width=0.28\textwidth]{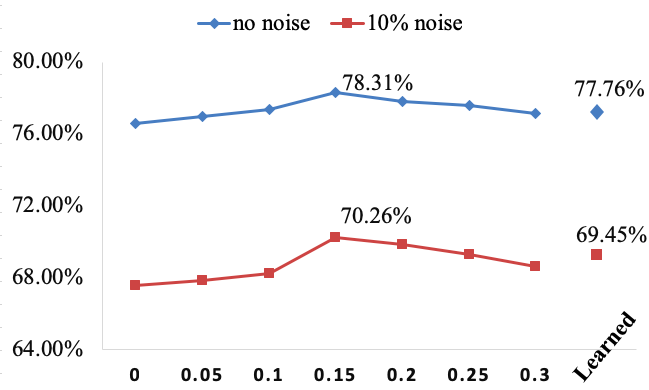}
%}
%\subfigure[$\delta_2$]{
\includegraphics[width=0.28\textwidth]{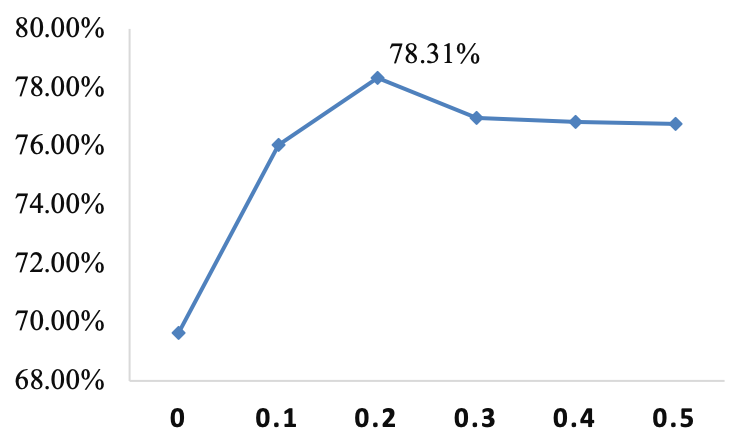}
%}
%\subfigure[$\beta$]{
\includegraphics[width=0.28\textwidth]{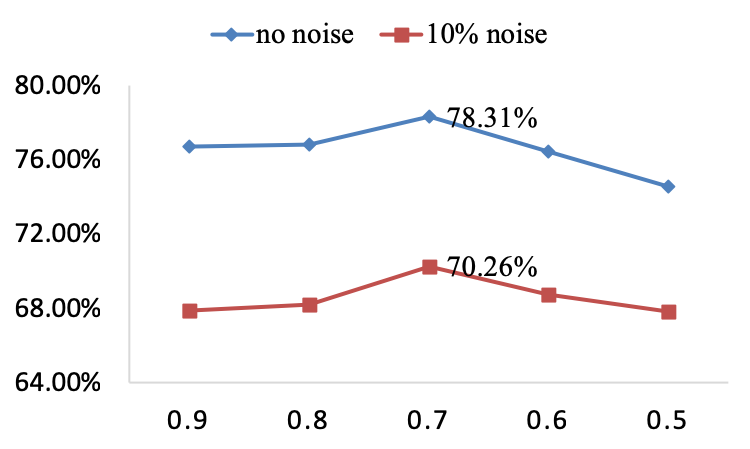}
%}
	\caption{Evaluation of the margin $\delta_1$ and $\delta_2$, and the ratio $\beta$ on the RAF-DB dataset.}
	\label{fig:phi}
\end{figure*}

\begin{table}[t!]
\center
\setlength{\abovecaptionskip}{0.cm}
\caption{Evaluation of the ratio $\gamma$ between RR-Loss and WCE-Loss.}
\begin{tabular}{@{}cccccccccc@{}}
\toprule
%Ratio  
%& 1:5  & 1:2 & 1:1  &2:1  & 5:1 \\ \midrule
& 0.2  & 0.3 & 0.5  & 0.6  & 0.8 \\ \midrule
%Performance (\%) 
& 76.12\%  & 76.35\% & \textbf{78.31\%}  &76.57\%  & 71.75\% \\ 
\bottomrule
\end{tabular}
\label{tab:ratio}
\end{table}

\begin{table*}[htp]
\center
\setlength{\abovecaptionskip}{0.cm}
\small
\caption{Comparison to the state-of-the-art results.$^*$These results are trained using label distributions. $^+$Oversampling is used since AffectNet is imbalanced. $^\ddagger$RAF-DB and AffectNet are jointly used for training. Note that IPA2LT tests with 7 classes on AffectNet.}\label{tab:soa}
\begin{subtable}{.3\linewidth}\centering
\setlength{\abovecaptionskip}{0.cm}
\caption{Comparison on RAF-DB.}\label{tab:rafdb}
{\begin{tabular}{@{}cc@{}}
\toprule
Method  & Acc. \\ \midrule
DLP-CNN \cite{li2017reliable}  & 84.22\\
%IPA2LT-EM \cite{zeng2018facial} & 85.30 \\
IPA2LT \cite{zeng2018facial} & 86.77 \\
gaCNN \cite{li2018occlusion} & 85.07 \\
RAN \cite{wang2019region} & 86.90 \\ \midrule
Our SCN (ResNet18) & \textbf{87.03} \\
Our SCN (ResNet18) $^\ddagger$& \textbf{88.14} \\
\bottomrule
\end{tabular}
}
\end{subtable}%
~
\begin{subtable}{.3\linewidth}\centering
\setlength{\abovecaptionskip}{0.cm}
\caption{Comparison on AffectNet.}\label{tab:affectnet}
{\begin{tabular}{@{}cc@{}}
\toprule
Method  & mean Acc. \\ \midrule
Upsample \cite{mollahosseini2017affectnet}  & 47.00 \\
Weighted loss \cite{mollahosseini2017affectnet}  & 58.00 \\
IPA2LT$^\ddagger$ \cite{zeng2018facial} (7 cls) & 55.71 \\
RAN \cite{wang2019region} & 52.97 \\
RAN$^+$ \cite{wang2019region} & 59.5 \\ \midrule
Our SCN$^+$(ResNet18) & \textbf{60.23} \\
\bottomrule
\end{tabular}}
\end{subtable}
~
\begin{subtable}{.3\linewidth}\centering
\setlength{\abovecaptionskip}{0.cm}
\caption{Comparison on FERPlus}\label{tab:ferplus}
{\begin{tabular}{@{}cc@{}}
\toprule
Method  & Acc. \\ \midrule
PLD$^*$ \cite{barsoum2016training}  & 85.1 \\
ResNet+VGG \cite{huang2017combining}  & 87.4 \\
SeNet50$^*$ \cite{albanie2018emotion} & 88.8 \\
RAN \cite{wang2019region} & 88.55 \\
RAN-VGG16$^*$ \cite{wang2019region} & 89.16 \\ \midrule
Our SCN (ResNet18/IR50) & 88.01/\textbf{89.35} \\
\bottomrule
\end{tabular}}
\end{subtable}
\end{table*}

\subsection{Ablation Studies}
\textbf{Evaluation of the three modules in SCN}.  %\pxj{Table \ref{tab:modules}}
To evaluate the effect of each module of SCN, we design an ablation study to investigate WCE-Loss, RR-Loss and Relabel modules on RAF-DB. We show the experimental results in Table \ref{tab:modules}. %To make a fair comparison, we validate the SCN with direct train setting. 
\pxj{
Several observations can be concluded in the following. First, for both training schemes, a naive relabeling module (2nd row) added into the baseline (1st row) can degrade performance slightly. This may be explained by that many relabeling operations are wrong from the baseline model. It indirectly indicates that our elaborately-designed relabeling in the low-importance group with rank regularization is more effective. Second, when adding one module, we obtain the highest improvement by WCE-Loss which improves the baseline from 72\% to 76.26\% on RAF-DB. This suggests that the re-weighting is the most contributed module for our SCN. Third, the RR-Loss and the relabeling module can further boost WCE-Loss by 2.15\%.
}
%We first show the results that training without WCE-Loss, RR-Loss, and Relabel modules. Then, we gradually add those three modules into training stage. When we only add one module into the training. WCE-Loss improves 6.26\%, 0.45\%, and 0.42\% on three dataset. RR-Loss improves 2.15\%, 0.17\%, and 0.25\% on three datasets. Specially, only adding relabel module into training do not improve the performance, because the CNNs can not find the annotation noises and relabel them without WCE-Loss and RR-Loss. When WCE-Loss and RR-Loss jointly optimize the CNNs, the performance is improved by 4.57\%, 0.60\%, and 0.67\% on three datasets. The last line results from Table \ref{tab:modules} show that adopting three modules can dramatically improve the FER performance with 6.31\%, 1.00\%, and 1.02\% on three datasets. Those results help us know the effect of the three modules and how to use them to boost FER performance according to the situations. 
%In general, WCE-Loss is used to weight the samples by importance values, which can reduce the influence of the noisy annotations. RR-Loss ensure a margin between the high and low importance group and offer a reference for Relabel module. Relabel module is used to find the noisy annotations and try to `cure' those samples by itself.

\textbf{Evaluation of the ratio $\gamma$}. %\pxj{Table \ref{tab:ratio}}
% What does the ratio mean? use one sentence.
\pxj{In Table \ref{tab:ratio}, we evaluate the effect of different ratios between the RR-Loss and WCE-Loss. We find that setting equal weight for each loss achieves the best results.
Increasing the weight of RR-Loss from 0.5 to 0.8 dramatically degrades performance which suggests that WCE-Loss is more important.}
%In Table \ref{tab:ratio}, we set 5 different ratio and show the recognition performance on the RAF-DB dataset. We find the 1:1 is the best ratio and too small ratio of WCE-Loss leads to a drop of performance.

\textbf{Evaluation of $\delta_1$ and $\delta_2$}.  %\pxj{Figure}.
$\delta_1$ is a margin parameter to control the mean margin between the high- and low-importance groups. For fixed setting, we evaluate it from 0 to 0.30.  Figure \ref{fig:phi} (left) shows the results for both fixed and learned $\delta_1$. The default $\delta_1$ = 0.15 obtains the best performance, which shows that the margin should be an appropriate value. 
We also design a learnable paradigm of $\delta_1$, and initialize it to 0.15. The learnable $\delta_1$ converges to $0.142\pm0.05$ and the performances are 77.76\% and 69.45\% in original and noise RAF-DB datasets, respectively.

%\textbf{Evaluation of $\delta_2$}
$\delta_2$ is a margin to determine when to relabel a sample. The default $\delta_2$ is 0.2. We evaluate $\delta_2$ from 0 to 0.5 on original RAF-DB, and show the results in Figure \ref{fig:phi} (middle). $\delta_2=0$ means we relabel a sample if the max prediction probability is larger than the probability of the given label. Small $\delta_2$ leads to a lot of incorrect relabeling operations which may hurt performance significantly. Large $\delta_2$ leads to few relabeling operations which converges to no relabeling. We get the best performance in 0.2.
% From Table \ref{tab:modules}, we can see the relabel module further improves performance on the three target datasets. We evaluate the parameter $\delta_2$ in Figure \ref{fig:phi}. Increasing $\delta_2$ from 0 to 1.0 gradually improves the performance while larger $\delta_2$ leads to degradation to the baseline performance. The reason may be that the smaller $\delta_2$ may lead the relabel too easily and the too bigger $\delta_2$ may make the relabel module out of action.

\textbf{Evaluation of the $\beta$}.
$\beta$ is the ratio of high importance samples in a minibatch. We study different ratios from 0.9 to 0.5 in both synthetic noisy and original RAF-DB dataset. The results are shown in Figure \ref{fig:phi} (right). Our default ratio is 0.7 that achieves the best performance. 
Large $\beta$ degrades the ability of SCN since it considers few of the data is uncertain. Small $\beta$ leads to over-consideration of uncertainties which decreases the training loss unreasonably.

%Table \ref{fig:phi} shows the results with the different ratios. Although the ratio changes a lot, the accuracy of the SCN remains in a stable range. Even the ratio is 0.5, the results are improved by 2.57\% and 6.37\% compared with baseline methods. This indicates that $\beta$ is not a sensitive hyper-parameter.
% It demonstrates that our SCN suffers from the major changes of the ratio.
%the ratio is not a sensitive parameter to the performance.

\subsection{Comparison to the State of the Art}

Table \ref{tab:soa} compares our method to several state-of-the-art methods on RAF-DB, AffectNet, and FERPlus.  %deng2019arcface
IPA2LT~\cite{zeng2018facial} introduces the latent ground-truth idea for training with inconsistent annotations across different FER datasets. gaCNN \cite{li2018occlusion} leverages a patch-based attention network and a global network. RAN\cite{wang2019region} utilizes face regions and original face with a cascade attention network. gaCNN and RAN are time-consuming due to the cropped patches and regions. Our proposed SCN does not increase any cost in inference. Our SCN outperforms these recent state-of-the-art methods with \textbf{88.14\%}, \textbf{60.23\%}, and \textbf{89.35\%} (with IR50~\cite{deng2019arcface}) on RAF-DB, AffectNet, and FERPlus, respectively.

%% file: Conclusion.tex
\pxj{ 
This paper presents a self-cure network (SCN) to suppress the uncertainties of facial expression data thus to learn robust feature for FER. The SCN consists of three novel modules including self-attention importance weighting, ranking regularization, and relabeling. The first module learns a weight for each facial image with self-attention to capture the sample importance for training and is used for loss weighting. The ranking regularization ensures that the first module learns meaningful weights to highlight certain samples and suppress uncertain samples. The relabeling module attempts to identify mislabeled samples and modify their labels. Extensive experiments on three public datasets and our collected WebEmotion show that our SCN achieves state-of-the-art results and can handle both synthetic and real-world uncertainties effectively.}
%Moving forward, we will further study more effective label noise suppression methods, e.g. by introducing iterative sample relabelling.}

%This paper proposes the self-cure network for robust feature learning of FER under label noises. SCN mainly includes three modules, namely self-attention noise weighting, ranking regularization, and noise relabeling. The first module learns a weight for each face with self-attention to capture the sample importance for training and is used for loss weighting. The ranking regularization ensures that the first module learns meaningful weights to highlight accurate annotations and to weaken wrong annotations. The last relabeling module aims to relabel these samples with low-importance weights. Experiments on synthetic noises and our collected WebEmotion data validate the effectiveness of our method. The SCN significantly improves vanilla CNNs which do not consider annotation noises. We achieve state-of-the-art results on three widely-used FER datasets.}